\documentclass{bmvc2k}
\usepackage{lipsum}
\usepackage{framed,multirow}
\usepackage{booktabs}
\usepackage{url}
\usepackage{xcolor}

\usepackage{comment}

\usepackage{algorithm}
\usepackage[noend]{algorithmic}
\usepackage{eqparbox}

\usepackage{graphicx}
\usepackage{amsmath,amssymb} %
\usepackage{bm}
\usepackage{booktabs}
\usepackage{rotating,multirow}
\usepackage{wrapfig}

\usepackage{tikz}
\usepackage{subfigure}
\usepackage[export]{adjustbox}

\def\eg{\emph{e.g}\bmvaOneDot} 
\def\ie{\emph{i.e}\bmvaOneDot} 
 
\def\etc{\emph{etc}\bmvaOneDot} 
 
\def\etal{\emph{et al}\bmvaOneDot}
\makeatother

\graphicspath{{./Images/}{./Figures/result/}{./Images/result/}}

\newcommand{\ua}{$\uparrow$}
\newcommand{\da}{$\downarrow$}
\renewcommand{\bf}{\bfseries}

\newcommand{\hleft}{%
\begin{tikzpicture}
\draw[fill] (0, 0) -- (0ex, 0ex) arc(90:270:1ex) -- (0, 0);
\end{tikzpicture}%
}
\newcommand{\hleftk}{%
\begin{tikzpicture}
\draw[densely dotted] (0, 0) arc(90:270:1ex) -- (0, 0)  ; \draw[fill]  (0, 0) arc(90:180:1ex) -- (0, -1ex) -- (0, 0);
\end{tikzpicture}%
}
\newcommand{\hright}{%
\begin{tikzpicture}
\draw[fill] (0, 0) -- (0ex, 0ex) arc(-90:90:1ex) -- (0, 0);
\end{tikzpicture}%
}
\newcommand{\hrightk}{%
\begin{tikzpicture}
\draw[densely dotted] (0, 0) arc(-90:90:1ex) -- (0, 0);
\draw[fill](0, 1ex) -- (1ex, 1ex) arc(0:90:1ex); 
\end{tikzpicture}%
}

\title{Toward multi-task learning for object detection and semantic segmentation on partially annotated data}
\title{One task at a time: toward combining \\ detection and segmentation \\ on partially annotated data}
\title{Data exploitation: combing \\ detection and segmentation \\ on partially annotated data}
\title{Data exploitation: learning both \\ object detection and semantic segmentation \\ on partially annotated data}
\title{Data exploitation: multi-task learning of \\ object detection and semantic segmentation \\ on partially annotated data}

\addauthor{Hoàng-Ân Lê}{hoang-an.le@irisa.fr}{}
\addauthor{Minh-Tan Pham}{minh-tan.pham@irisa.fr}{}

\addinstitution{
IRISA, Université Bretagne Sud, 
\\
UMR 6074, 56000 Vannes, France 
}

\runninghead{H.-Â. Lê and M.-T. Pham}{Data exploitation with multi-task learning}

\begin{document}

\maketitle

\begin{abstract}
Multi-task partially annotated data where each data point is annotated for
only a single task are potentially helpful for data scarcity
if a network can leverage the inter-task relationship.
In this paper, we study the joint learning of object detection
and semantic segmentation, the two most popular vision problems,
from multi-task data with partial annotations.
Extensive experiments are performed to evaluate each
task performance and explore their complementarity when
a multi-task network cannot optimize both tasks simultaneously.
We propose employing knowledge distillation to leverage
joint-task optimization. The experimental results show favorable
results for multi-task learning and knowledge distillation over
single-task learning and even full supervision scenario.
All code and data splits are available at
\url{https://github.com/lhoangan/multas}
\end{abstract}

\section{Introduction}
\label{sec:intro}

Although both object detection and semantic segmentation
aim to understand the image content, the two problems differ in
spatial structure and information granularity.
Object detection performs at the object level
outputting unordered list of bounding boxes
with corner coordinates and object types
while semantic segmentation provides per-pixel predictions; object detection
distinguishes object instances while semantic segmentation
recognizes each category as a whole and also amorphous
regions such as ground, sky, sea,~\etc.

Attempts have been made to jointly learn both tasks in a single model.
Methods such as Mask R-CNN~\cite{He2017} overcomes the spatial structure difference by
generating an object mask for each predicted bounding box, effectively
predicting instance segmentation. On the other hand, the introduction of
panoptic segmentation~\cite{Kirillov2019panopticFPN} can be seen as resolving
the information granularity difference in which instance-level
objects and amorphous categories are tackled together as a dense prediction
problem.
Combining both tasks under the common form of instance segmentation, however,
leaves the original tasks unfinished: Mask R-CNN does not provide
segmentation masks for stuff categories nor does panoptic segmentation directly
provide bounding box coordinates.

Multi-task learning is a research area that allows training different problems
under the same model. The general assumption is that several tasks are inherently
related to one another and by optimizing them together for each input image,
the network could extract common features and pick up the salient interrelationships.
Although training multiple tasks could potentially increase tasks coherency and,
for particular setups, also allow self-supervision~\cite{Casser2019struct2depth,Zhou2017DepthEgo},
it is challenging as each task would require specific
architecture and optimization criteria, and maintaining a training dataset
with consistent annotations for all tasks proves to be expensive.

\begin{figure}
    \centering
    \includegraphics[width=.32\textwidth]{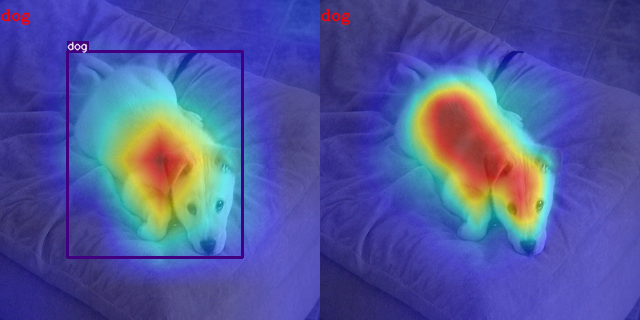}
    \includegraphics[width=.32\textwidth]{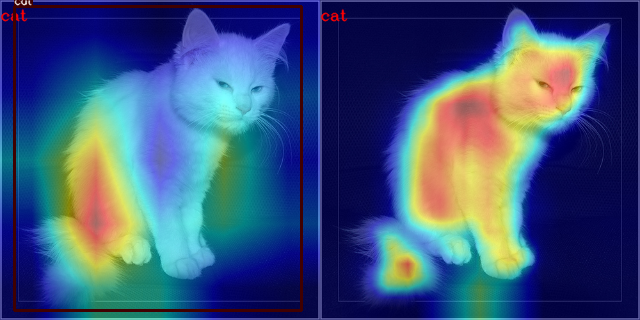}
    \includegraphics[width=.32\textwidth]{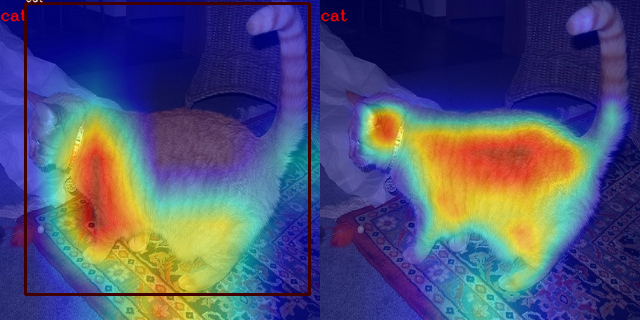}
    \caption{
    Class activation maps at the same feature layers of an object detection
    network (left) and semantic segmentation (right) showing incompatible feature
    attentions: detection only activates
    a (few) feature at the scale producing fit
    boxes while segmentation activates all those belong
    to the objects.
    }
    \label{fig:CAM1}
\end{figure}

In this paper, the joint learning of object detection and semantic segmentation
is considered, which despite their popularity as single tasks,
seems to receive limited attentions in the literature.
Due to different targets, although the two tasks 
are closely related, the features learned for each task are not readily compatible.
Figure~\ref{fig:CAM1} shows the activation map using Grad-CAM~\cite{Selvaraju2017gradcam}
at the same layers of two networks with the same encoder architectures,
trained for object detection and semantic segmentation. Semantic segmentation activates
(nearly) all the features covering the object of interest while object detection activates
only those at the feature scale that produces fitted bounding boxes, no matter if they
belong to the objects. Table~\ref{tab:single_multiple} and Sec.~\ref{sec:single_multi_task}
show that an encoder trained for one task cannot immediately be used for the
other tasks when only the task-specific head is finetuned.

Diverging from the usual multi-task learning assumption that
annotations are available in all tasks
for each training example, we limit the scope of the paper
to multi-task partially annotated data, where each image is annotated for a single
task and there are no images containing both task annotations.
This is interesting because
(1) the network cannot optimize both tasks for the same input and is hindered
in attempt to learn joint features and salient interrelationships;
(2) therefore, this setting would illustrate the complementarity of the two
tasks of interest;
and (3) it is data efficient and would be an alternative method to ameliorate 
the data scarcity problem as more data with single-task
annotations could be used for training, allowing for expanding training ability.

To that end, we employ a simple multi-task learning framework to study the %
combination of object detection and semantic segmentation.
We experiment with various setups and observe each task's performance
under different input conditions.
By varying the datasets, the interaction
between the two tasks can be observed which can be useful for further
study. The simple feature-imitation knowledge distillation model is employed
for cross-task optimization which is seemingly not possible for partially annotated data.

The paper contributions are as follows. We explore the combination of object
detection and semantic segmentation in a multi-task learning framework for
partially annotated data. Extensive experiments are performed to evaluate
both quantitatively and qualitatively the benefit of one task to the other.
A knowledge distillation method is employed and evaluated for joint-task
optimization.

\section{Related work}
\label{sec:relatedwork}

\subsection{Multi-task learning}

Multi-task learning (MTL) trains a single model that can infer
different task targets from a given input. One of the main assumptions
is the compatibility of the features learned for each task and
by optimizing them for each input, the network could learn
the common knowledge that benefits and complements one
another~\cite{Lu2021taskology,Li2022Universal}.

Several methods have been proposed to accommodate various tasks
and network architectures to improve shared information among
the tasks~\cite{Bruggemann2021} using
attention mechanisms~\cite{Liu2019mtan}
and gating strategies~\cite{Bruggemann2020}, and to study
cross-task relationships~\cite{Lu2021taskology,Zamir2020xtaskConsist}.
The fully supervised learning strategy requires annotations
available for all tasks per training example for optimization,
which is costly and hinders scalability.
Therefore, attempts have been made
for semi-supervised learning~\cite{Chen2020meanTeacher,Imran2020}
that allows learning from unlabelled data and relaxes the number of annotations,
yet all-task annotations per training sample are still required.

Closely related to the problem in our paper is the work of 
Li~\etal~\cite{Li2022MTPSL}, in which each training data point is only
required to contain an annotation for a single task,
or the multi-task partial annotation scenario. The cross-task consistency
constraint is proposed and the task-specific annotations are projected to the joint
pairwise task-space from which supervised signals are provided to the training process.
The method requires the dense spatial structures of the annotations
making it inapplicable for object detection in this paper.

\subsection{Knowledge distillation}

Hinton~\textit{et al.}\cite{Hinton2015} has shown that a network could
benefit from a larger or an ensemble of models, called teachers,
by mimicking the predicted logits or imitate the deep features~\cite{Guo2021Defeat}
learned by them.
Depending on the purpose, knowledge distillation (KD) could be seen as model
compression which aims to reduce model complexity with less performance sacrificing,
or a self-training technique~\cite{Xie2020selftraining,Zoph2020ST} 
where a network is trained using the combination of
available annotations and pseudo-labels provided by the teacher's predictions.
Self-training with uncertain teachers for object detection
has recently been studied~\cite{Le2023iccw}, where
the teachers are trained with a small number of supervised data disjoint with
the students' training set, or for a different task (segmentation).
Different from their paper which also involves
detection-segmentation multi-task training but
focuses only on the detection benefit,
our work interests in both tasks' performances and shows
the multi-task advantage with TIDE~\cite{tide-eccv2020} error analysis.
Multi-task learning has seen other applications with self-training such as
the extension of Born Again Network~\cite{Furlanello2018BAN} for learning context in NLP
problem~\cite{Clark2019BAM} using a weight annealing
strategy to update the distillation and multi-task losses. Li~\etal~\cite{Li2020KDMT}
apply knowledge distillation to solve the unbalanced loss optimization problem in 
multi-task learning and show favorable results for
fully annotated semantic segmentation and depth prediction training.

\section{Method}
\label{sec:method}

\begin{figure}[t]
    \centering
     \def\svgwidth{.85\textwidth}
    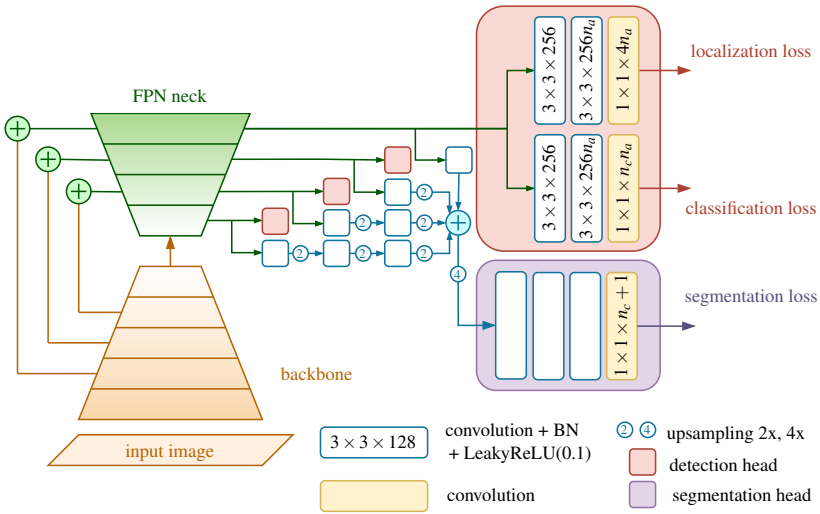
    \caption{A general view of the network architecture including an encoder (backbone and neck)
    and 2 heads for object detection and semantic segmentation. Object detection
    performs at each pyramid scale while semantic
    segmentation aggregates all scales and upsamples them to the image-size.
    }
    \label{fig:full_pipeline}
\end{figure}

To study the relationship between object detection and semantic segmentation,
we apply a simple multi-task learning framework following the
encoder-decoder principle. 
Common to many approaches is a shared encoder, comprising a backbone
(\eg~the ResNet family) and a neck (\eg~the FPN family),
which extracts and aggregates features from input images
while multiple decoders, or heads, provide task-specific predictions.
The overview framework is shown in Figure~\ref{fig:full_pipeline}.
In this work, the one-stage anchor-based object detection
architectures~\cite{liu2016ssd,Lin2017focal} are studied.
The detection head (red boxes) at each pyramid scale comprises 2 output branches with the same
architecture for localization and classification losses.
For semantic segmentation, the multi-scale pyramid features are aggregated using
the architecture by Kirillov~\etal~\cite{Kirillov2019panopticFPN}.
The aggregation is performed by alternating between convolution and double up-sampling
the features at each scale until one-fourth of the input size before element-wise adding
together and finally quadruple up-sampling to the input size. 
The aggregated and segmentation features' dimensions
are set to 128 following the original work while detection heads features are 256 as output
from the encoder.
The detection head uses the Focal Loss~\cite{Lin2017focal} and the Balanced L1 Loss~\cite{Pang2019}
for localization and classification while the semantic segmentation head uses the regular
cross-entropy with softmax loss.

\textbf{Multi-task training.}$\quad$
As each data point is annotated for only a single task, not all the losses can be optimized
together. Two optimization approaches are considered, alternating the tasks (1) every
epoch or (2) every iteration. For the former,
the network is trained with one task for one epoch with the gradients computed from the
respective task-specific head and leaving the other task head untouched
before being trained with the other task in the following epoch.
For the latter,
a mini-batch of images with annotations for one task
is passed through the network immediately after one with the other task.
The gradients from each mini-batch is computed for the corresponding head and
accumulated for the encoder. Only after mini-batches from both tasks have been
fed in and gradients accumulated are the network parameters updated.
As a result, both tasks start and end an epoch together. Thus, the task
with fewer annotations will randomly have some images repeated in waiting for
a new epoch. We show in Sec.~\ref{sec:single_multi_task}
and Table~\ref{tab:single_multiple} the performance difference of the two strategies.

\begin{figure}[t]
    \centering
    \def\svgwidth{.6\textwidth}
    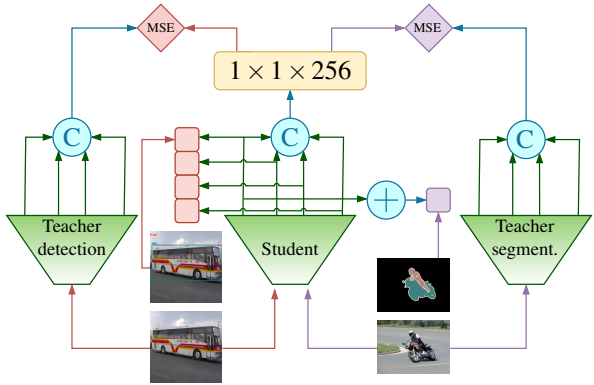
    \caption{Visualization the knowledge distillation process in the
    multi-task learning for partially annotated data. Images and task-specific
    annotations are fed to the student and the respective teachers,
    with KD losses computed on teacher-student flattened and concatenated neck features.
    }
    \label{fig:pipeline_KD}
\end{figure}

\textbf{Knowledge distillation.}$\quad$
We concatenate the features of all scale levels
along the flattened spatial dimensions. The features of the student network
are projected by a $1\times1$ convolution before being compared to the 
corresponding teachers'.
The simple Mean Square Error (MSE)~\cite{Zhang2021PDF} is applied for feature imitation distillation.
The illustration is shown in Figure~\ref{fig:pipeline_KD}.

As each training image can optimize a single task, there are 3 cases for distilling
the student features per iteration: (1) from the teacher whose task is annotated (1mse)
so the student's features are forced to follow the teacher's while learning from
the provided ground truths at the same time,
(2) from the task teacher \textit{without} annotations (0mse) so that the head is trained
with one task (using ground truth) while the encoder is forced to follow the other's
teacher, and (3) is the combination of both (2mse).

\section{Experiments}
\label{sec:exp}

\subsection{Setup}

\paragraph{Datasets.}
All the experiments are conducted on the Pascal VOC~\cite{PascalVOC} containing 20 object
categories with 8,218 bounding-box annotated images for training, 8,333 for validation,
and 4,952 for testing. Due to limited semantic segmentation annotations originally provided 
(1,464 and 1,449 for training and validation, respectively),
the common practices use extra annotations provided by~\cite{Hariharan2011SBD},
resulting in 10,582 training images.

To simulate the partial supervision scenario, images are randomly sampled into 2 subsets,
one for detection \hleft\ whose semantic annotations are held back and the other
for semantic segmentation \hright\ whose bounding-box annotations are kept out,
resulting in 7,558 and 7,656 respectively (images without semantic segmentation
ground truth are prioritized to the detection subset).
For validation, the originally provided validation set for semantic segmentation
with both task annotations are used with 1,443 images 
(6 images with only semantic segmentation are withheld).
Unless stated otherwise, the image lists are kept the same in all experiments.

\begin{table}[t]
    \centering
    \setlength{\tabcolsep}{2.8pt}
    \begin{tabular}{@{}llcc@{}}
        \toprule
        Training &                  & \multicolumn{2}{c}{Detection} \\
        \cmidrule{3-4}       
                                    &                &   RN18+FPN & RN50+PAFPN \\ %
        \midrule       
        Single task                 & \hleft         &      42.81 &     50.22  \\ %
        \midrule       
        Finetuning head             & \hleft~\hright &      31.80 &     36.08  \\ %
        Finetuning full             & \hleft~\hright &      43.30 &     50.61  \\ %
        ReCAM~\cite{Chen2022ReCAM}  & \hleft~\hright &            &            \\ %
        \midrule                             
        Multi-task (epoch)          & \hleft~\hright &      44.51 &     51.62  \\ %
        Multi-task (iteration)      & \hleft~\hright &  \bf 44.78 & \bf 52.10  \\ %
        \midrule                             
        \midrule                             
        Multi-task (full)          & \hleft~\hright &      46.83 &     53.68  \\ %
        \bottomrule
    \end{tabular}~
    \begin{tabular}{@{}lcc@{}}
        \toprule
                          & \multicolumn{2}{c}{Segmentation} \\
        \cmidrule{2-3}
                          & RN18+FPN  & RN50+PAFPN \\ %
        \midrule
                 \hright  &     64.55 &     72.32  \\ %
        \midrule
        \phantom{\hright} &     61.08 &     65.84  \\ %
        \phantom{\hright} &     65.21 &     72.01  \\ %
        \phantom{\hright} &     63.30 &     70.11  \\ %
        \midrule
        \phantom{\hright} &     66.99 &     72.93  \\ %
        \phantom{\hright} & \bf 67.57 & \bf 73.66  \\ %
        \midrule
        \midrule
        \phantom{\hright} &     67.47 &     73.08  \\ %
        \bottomrule
    \end{tabular}
    \caption{
    Comparing single tasks and multi-task learning on partially annotated data.
    Multi-task are trained by alternating the tasks every epoch
    or iteration.
    Training both tasks (full) results are included for reference.
    Multi-task learning outperforms all other settings.
    }
    \label{tab:single_multiple} %
\end{table}

\begin{table}[]
    \centering
    \setlength{\tabcolsep}{5.0pt}
         \begin{tabular}{@{}llcc@{}}
        \toprule
        Training                            &&  \multicolumn{2}{c}{Detection} \\
        \cmidrule(l){3-4}  
                                            &&  RN18+FPN & RN50+PAFPN \\
        \midrule
                 Single task &\hleft         &     42.81 &  50.22    \\ %
        \phantom{Single task}&\hleftk        &     38.10 &  43.73    \\ %
        \midrule
                 Multi-task &\hleft~\hright  &     44.78 & \bf 52.10 \\ %
        \phantom{Multi-task}&\hleftk~\hright &     40.89 &  47.27    \\ %
        \phantom{Multi-task}&\hleft~\hrightk & \bf 44.99 & 51.43     \\ %
        \bottomrule
    \end{tabular}
    \begin{tabular}{@{}ccc@{}}
         \toprule
         & \multicolumn{2}{c}{Segmentation} \\
         \cmidrule(r){2-3}
         & RN18+FPN & RN50+PAFPN \\
         \midrule
                  \hright   &     64.55 & 72.32     \\ %
                  \hrightk  &     63.02 & 68.96     \\ %
         \midrule
         \phantom{\hright } & \bf 67.57 & \bf 73.66 \\ %
         \phantom{\hright}  &     65.39 & 73.17     \\ %
         \phantom{\hrightk} &  66.16    & 73.03     \\ %
         \bottomrule
    \end{tabular}
    \caption{
    Single-task and multi-task performance when trained with half annotated
    detection \protect\hleftk\ and segmentation \protect\hrightk.
    Single-task is impacted more from the reduced training sizes
    for the respective task.
    }
    \label{tab:half_data} %
\end{table}

For out-of-domain experiments,
the Cityscapes~\cite{cityscapes2016} dataset with 2,975 training, 500 validation images,
and 7 semantic classes is employed for semantic segmentation.
We resize the images to $128\times256$ to speed up the training process
following~\cite{Li2022MTPSL,Liu2019mtan}.

\paragraph{Network architectures.}
For comparison purposes, two backbone models from the ResNet family are employed,
including ResNet50 backbone with PAFPN~\cite{Liu2018PAFPN} neck (RN50+PAFPN), and 
ResNet18 backbone with FPN~\cite{Lin2017} neck (RN18+FPN).
A few modifications are made following the implementation of~\cite{Zhang2021PDF},
including removing the first max-pooling layer of ResNet as in ScratchDet~\cite{Zhu2019scratchdet},
and adding the context enhancement module as in ThunderNet~\cite{Qin2019thundernet}.
The number of convolutional blocks in the detection head subsets is reduced from
4 to 2 to speed up the training time.
The two networks are also used for knowledge distillation as teacher and student,
respectively, with parameter ratio of 1.61. The networks are trained for 30 epochs,
with learning rate of $5\times10^{-3}$.

\paragraph{Evaluation and analysis.} We employ, for semantic segmentation
the conventional IOU score~\cite{Jaccard1912} and, for object detection,
the mAP metric implemented by the Detectron2 library~\cite{Wu2019detectron2},
which follows the original VOC code but averages APs at multiple IOU thresholds
in the range $[.5, .95, .05]$. The detection results at IOU of $50\%$,~\ie~AP50, are used
as inputs to the TIDE~\cite{tide-eccv2020} framework for analyzing the error sources.
TIDE breaks down detection errors into 6 types and estimates the isolated contribution
of each to the overall performance, as follows:
(1) \textbf{Cls} errors localize correctly but classify incorrectly;
(2) \textbf{Loc} errors classify correctly but localize incorrectly;
(3) \textbf{Both} errors classify and localize incorrectly;
(4) \textbf{Dupe} errors would be correct if not for a higher scoring detection;
(5) \textbf{Bkg} errors detect background as foreground;
(6) \textbf{Miss} errors are all undetected ground truths not already covered by Cls or Loc error.

\input{Tables/Figure3}

\subsection{Single-task learning and multi-task learning}
\label{sec:single_multi_task}

In this experiment, we confirm the benefit of multi-task training for
partially annotated data where each training example is only annotated 
for a single task. Possible data exploitation includes
(1) training a single-task network with the provided annotated data;
(2) training a single-task network by finetuning one pretrained for the other task;
(3) training a single-task network with provided annotated data and pseudo labels
generated by a semi-supervised learning method for the other task's data; and
(4) training a multi-task network with 2 decoders.
Except for (1), all other strategies involve the data from the other task
in the training process: (2) is a standard transfer learning approach and
(3) formulates as a weak-supervised problem. The ReCAM method~\cite{Chen2022ReCAM}
is applied to generate semantic mask for images from the detection subset \hleft.
For the transfer learning approach, we also include experiments where the
pretrained backbone and neck are kept frozen during finetuning to show the compatibility
of the features extracted for one task to the other.
The results are shown in Table~\ref{tab:single_multiple}.

\begin{table}[t]
    \centering
    \begin{tabular}{@{}crrrrrrrrr@{}}
    \toprule
                     & AP50\ua  & Cls\da & Loc\da& Both\da& Dupe\da& Bkg\da& Miss\da& FP\da  &  FN \da  \\
    \midrule
    \hleftk~\hright  &   40.89 &    4.23&   7.16&   0.70 &\bf 0.37&\bf1.45&   8.38 &\bf13.12&   16.58    \\
    \hleft~\hrightk  &\bf44.99 &\bf 3.75&\bf6.08&\bf0.58 &    0.43&   1.46&\bf6.54 &   14.02&\bf13.09    \\
    $\Delta$         &    4.10 &   -0.48&  -1.08&  -0.12 &    0.06&   0.01&  -1.84 &   0.90 &   -3.49    \\
    \midrule
    \hleftk~\hright  &   47.27 &   2.69 &   6.81&   0.56 &   0.37 & 1.07  &   8.57 &   9.55 &   15.12    \\
    \hleft~\hrightk  &\bf51.43 &\bf2.46 &\bf6.06&\bf0.51 &\bf0.31 & 1.07  &\bf6.39 &\bf9.47 &\bf12.30    \\
    $\Delta$         &    4.16 &  -0.23 &  -0.75&  -0.05 &  -0.06 & 0     &  -2.18 &  -0.08 &   -2.82    \\
    \bottomrule
\end{tabular}
    \caption{TIDE analysis of detection from RN18+FPN (top) and RN50+PAFPN (bottom).}
    \label{tab:tide_half}
\end{table}

\begin{table}[t]
    \centering
    \setlength{\tabcolsep}{5.0pt}
\begin{tabular}{@{}llrrrrrrrrr@{}}
    \toprule
                      & & AP50\ua  & Cls\da & Loc\da& Both\da& Dupe\da& Bkg\da& Miss\da& FP\da  &  FN \da  \\
    \midrule
    STL&\hleft          &\bf42.81&    4.71&\bf6.17&   0.73 &   0.40 &   1.61&\bf6.70 &   16.40&\bf12.85  \\
    MTL&\hleftk~\hright &   40.89&\bf 4.23&   7.16&\bf0.70 &\bf0.37 &\bf1.45&   8.38 &\bf13.12&   16.58  \\
    $\Delta$           &&  -1.92 & -0.48  & 0.99  &  -0.03 &  -0.03 &  -0.16&  1.68  &  -3.28 &   3.73   \\
    \midrule
    STL&\hleft          &\bf50.22&   2.91 &\bf6.45&\bf0.54 &\bf0.34 &   1.25&\bf6.48 &  10.51 &\bf13.21  \\
    MTL&\hleftk~\hright &   47.27&\bf2.69 &   6.81&   0.56 &   0.37 &\bf1.07&    8.57&\bf9.55 &   15.12  \\
    $\Delta$           &&  -2.95 &  -0.22 &  0.36 &   0.02 &  0.03  & -0.18 &  2.09  &  -0.96 &   1.91   \\
    \bottomrule
\end{tabular}
    \caption{TIDE analysis of detection from RN18+FPN (top) and RN50+PAFPN (bottom).}
    \label{tab:tide_full}
\end{table}

It could be seen that the performance of multi-task learning even with partially
annotated data are highest for all settings. The results when the two tasks
are alternated every iteration seems to perform slightly better than every epoch.
We also include the results when an image is annotated with both task,~\ie fully
supervised setting with all annotations from~\cite{Hariharan2011SBD} for reference
(the results are not comparable as there are 10,476 images with both
task annotations). Even with fewer effective images, optimizing both tasks shows
superior performance.
Training a semantic network with joint annotated data and pseudo-label generated
by ReCAM method does not help even when compared to single task learning. 

Regarding transfer learning scenario, finetuning the whole network pretrained with the other task improves
over single-task learning while finetuning with frozen encoders plunges, showing
the incompatible features learned by one task to the other. Class activation maps
generated by Grad-CAM~\cite{Selvaraju2017gradcam} are shown in Figure~\ref{fig:gradcam_multi}
for single and multi-task networks at the same layer output by the encoder.
The multi-task activation seems spreading out for both detection and segmentation and
could recover a miss-detected object.

\paragraph{Multi-task with fewer data} We extend the study by adjusting the number of
training data used for multi-task learning. Half number of images from one task subset
are randomly removed while retaining those of the other. The results are shown in
Table~\ref{tab:half_data}.

Although reducing training data size takes a great toll on the 
respective single task performance, the multi-task results seem to have less impact,
especially for ResNet50+PAFPN with semantic segmentation task.
From a multi-task point of view, reducing detection set also affects
semantic segmentation performance, especially for RN18+FPN, while reducing
segmentation data does not seem to affect the performance of the detection counterpart.

TIDE analysis on Table~\ref{tab:tide_half} shows that the most contribution to the
difference between \hleftk~\hright\ and \hleft~\hrightk\ is FN, especially the 
missing objects (Miss, $\Delta$=-1.84, -2.18),
then faulty localization with correct classification (Loc, $\Delta$=-1.08, -0.75).
The Classification error (Cls) seems to be affected at a lesser degree ($\Delta$=-0.48, -0.23).
As the semantic segmentation masks do not contain precise locations
of object instances, even when VOC has limited instances per image,
using semantic masks helps more with classification and less
with localization. This is confirmed in Table~\ref{tab:tide_full}
where STL trained with full detection data~\hleft\ is compared to
MTL with half~\hleftk~\hright.
Although inferior in general performance, MTL with half detection
has lower classification error (Cls,$\Delta$=-0.48,-0.22)
background confusion (Bkg,$\Delta$=-0.16,-0.18), and generally
FP ($\Delta$=-3.28,-0.96).

\paragraph{Multi-task with different category sets}
To understand the interrelationships between object detection and semantic segmentation,
we gradually deviate one task from the other and observe the performance differences.
In this experiment, the category set for semantic segmentation is modified.
Various VOC semantic categories are merged into an ``abstract'' class
representing the group of the original labels such as the
\textit{vehicle} group (from aeroplane, bicycle, boat, bus, car, motorbike, and train),
\textit{animal} (from bird, cat, cow, dog, horse, and sheep),
\textit{furniture} (bottle, chair, dining table, potted plant, sofa, and TV monitor),
and \textit{person} as a group in itself.

Arranging different classes into the same group arrives at a
semantic segmentation task that aims to learn entirely different concepts from
the object detection task. The results are shown in Table~\ref{tab:diff_label_set}.
Although the performance distance between multi-task and single-task is shortened
as each task has to cope with its own concept targets, multi-task still has
its superiority. There is a diminishing return with higher capacity
architectures.

\paragraph{Multi-task with out-of-domain data}
In this experiment, the two tasks are further pushed to different data domains.
To that end, the semantic segmentation images are taken from the Cityscapes
dataset~\cite{cityscapes2016} with 7 classes.
Some of the classes are shared between the two datasets, such as car, human, vegetation.
The results in Table~\ref{tab:out_of_domain} show that jointly learning data from
different domains worsens the multi-task performance. It is not surprising
as the data belong to different distributions with different semantic concept targets,
the jointly learned features from one task is not helping but impede the other's
learned features.

\begin{table}[t]
    \centering
    \begin{tabular}{@{}lcc@{}}
     \toprule
     & \multicolumn{2}{c}{Detection (20 classes)} \\
     \cmidrule(r){2-3}
                        &      RN18+FPN  & RN50+PAFPN \\
     \midrule
     Single task           &     42.81   &     50.22 \\ %
     Multi-task            &  \bf 44.48  & \bf 50.38 \\ %
     \bottomrule
\end{tabular}
\begin{tabular}{@{}cc@{}}
     \toprule
     \multicolumn{2}{c}{Segmentation (4 classes)} \\
     \cmidrule(r){1-2}
     RN18+FPN & RN50+PAFPN \\
     \midrule
         78.47 &     81.82 \\ %
     \bf 79.32 & \bf 81.89 \\ %
     \bottomrule
\end{tabular}
    \caption{
    Single-task and multi-task performance when the tasks have
    different label sets. The performance gap
    decreases yet still in favor of multi-task learning.
    }
    \label{tab:diff_label_set} %
\end{table}

\begin{table}[t]
    \centering
    \begin{tabular}{@{}lcc@{}}
     \toprule
     &  \multicolumn{2}{c}{Detection (VOC)} \\
     \cmidrule(r){2-3}
     &  RN18+FPN & RN50+FPN \\
     \midrule
     Single task         & \bf 38.688 & \bf 44.683 \\ %
     Multitask           &     37.531 &     39.910 \\ %
     \bottomrule
\end{tabular}
\begin{tabular}{@{}cc@{}}
     \toprule
     \multicolumn{2}{c}{Segmentation (Cityscapes)} \\
     \cmidrule(r){1-2}
     RN18+FPN & RN50+FPN \\
     \midrule
     \bf 71.389  & \bf 72.398 \\ %
     69.481      &      70.247 \\ %
     \bottomrule
\end{tabular}
    \caption{
    Single tasks and multi-task performance when the tasks
    are from different domains. Jointly learning 2 tasks from different domains
    does not help but hurt the performance.
    }
    \label{tab:out_of_domain} %
\end{table}

\begin{table}[t]
    \centering
    \begin{tabular}{@{}llccc@{}}
    \toprule
    Training                       &                 &   Detection   &          & Segmentation \\
    \midrule    
             Single task           & \hleft          &      42.907   & \hright  &       65.291 \\ %
    \phantom{Single task} + KD     & \hleft          &      44.982   & \hright  &       67.375 \\ %
    \midrule   
             Multi-task            & \hleft~\hright  &      45.678   &          &       67.310 \\ %
    \phantom{Multi-task} + 1mse    & \hleft~\hright  &      45.989   &          &       69.126 \\ %
    \phantom{Multi-task} + 0mse    & \hleft~\hright  &      47.337   &          & \bf   70.056 \\ %
    \phantom{Multi-task} + 2mse    & \hleft~\hright  & \bf  47.611   &          &       69.911 \\ %
    \bottomrule
\end{tabular}
    \caption{
    Adding feature imitation knowledge distillation. For multi-task learning, the
    distilled features can be on the task with (1mse), or without annotations (0mse), or both (2mse).
    The performances are in favor for (0mse) and (2mse).
    }
    \label{tab:KD_singletask} %
\end{table}

\begin{table}[t]
    \centering
    \setlength{\tabcolsep}{5.0pt}
\begin{tabular}{@{}lrrrrrrrrrr@{}}
    \toprule
                             && AP50\ua  & Cls\da & Loc\da& Both\da& Dupe\da& Bkg\da& Miss\da& FP\da   &  FN \da  \\
    \midrule                           
    STK&\hleft                &   42.81 &\bf4.71 &   6.17&   0.73 &\bf0.40 &\bf1.61&   6.70 &\bf16.40 &   12.85  \\
    $\quad$+MSE&\hleft        &\bf44.98 &   4.94 &\bf5.70&\bf0.67 &   0.43 &   1.71&\bf5.39 &   17.32 &\bf11.67  \\
    $\Delta$                 &&  2.17   &   0.23 & -0.47 &  -0.06 &  0.03  &   0.10&   -1.31&    0.92 &   -1.18  \\
    \midrule                                      
    MTL &\hleft~\hright       & 44.78   &   3.70 &   6.72&   0.73 &\bf0.40 &   1.61&   6.13 &   14.90 &   12.43  \\
    $\quad$+MSE&\hleft~\hright&\bf47.61 &\bf3.10 &\bf6.17&\bf 0.65&   0.50 &   1.61&\bf5.56 &\bf13.47 &\bf11.66 \\
    $\Delta$                 &&  2.83   &  -0.60 & -0.55 &  -0.08 &  0.1   &   0   &   -0.57&   -1.43 &   -0.77  \\
    \bottomrule
\end{tabular}

    \caption{TIDE analysis of detection results with (+MSE) and without knowledge distillation.}
    \label{tab:tide_kd}
\end{table}

\subsection{Knowledge distillation}

In this section, knowledge distillation (KD) is used to enact joint-task training for
partially annotated data and multi-task learning.
To that end, the ResNet50+PAFPN architecture is used as the teacher model
and results of the student ResNet18+FPN are reported. The task-specific
heads are kept the same for the two networks. The teacher-student parameter ratio is 1.61.
Unless stated otherwise, the teachers are initialized with the corresponding
weights for single tasks from the previous experiments and stay frozen during
the training of the students.
The results are shown in Table~\ref{tab:KD_singletask}.
By simply forcing the student encoders to imitate the output of the teachers,
the corresponding results are improved, confirming the benefit of knowledge distillation.
Distilling the encoder neck features using one task's teacher while training
the other task's head using provided ground truths (0mse) shows favorable results
over distilling the same task that has annotations (1mse).
The results are even higher when both tasks are optimized simultaneously
in the fully-supervised scenario in Table~\ref{tab:single_multiple},
showing the benefit of multi-task data exploitation and
joint-task optimization using knowledge distillation.

Table~\ref{tab:tide_kd} shows the
errors reduced by KD for both single-task learning (STL) and multi-task learning (MTL).
It could be seen that among the first 6 errors, KD helps the most with Miss
detection and Cls. The effects, however, are not the same for STL and MTL: STL benefits
substantially from Miss error ($\Delta$=-1.31), reflecting also in FN ($\Delta$=-1.18)
while MTL benefits equally on Cls, Loc, and Miss, emphasizing on FP ($\Delta$=-1.43).
This suggests the more balancing performance of MTL over STL and the improvement of
robustness by KD.

\section{Conclusion}
\label{sec:conclusion}

The paper studies the possibility for jointly learning
object detection and semantic segmentation using 
partially annotated data. As there are no images 
with both task annotations, optimization is
alternated between the tasks. The experiments show that
by alternating every iteration, the networks could pick
up useful information from the other task's data
and improve over the single-task cases. Knowledge
distillation could be an alternative method allowing to learn 
interrelationship between one task and the other's features.

\section*{Acknowledgments}
This work was supported by the SAD 2021 ROMMEO project (ID 21007759) and the ANR AI chair OTTOPIA project (ANR-20-CHIA-0030).

\bibliography{macro, IRISA-BMVC23.bib}
\end{document}